\documentclass[review]{elsarticle}

\usepackage{lineno,hyperref}
\modulolinenumbers[5]

\journal{Neural Networks}

\include{def}

\usepackage{color}
\usepackage{graphicx}
\usepackage{amsmath}
\usepackage{booktabs}
\usepackage{algorithm}
\usepackage{algorithmic}
\usepackage{multirow}

\usepackage{booktabs}       
\usepackage{amsfonts}       
\usepackage{nicefrac}       
\usepackage{microtype}      
\usepackage{multirow}
\usepackage{graphicx}
\usepackage{amsmath}
\usepackage{arydshln}

\newcommand{\Rmnum}[1]{\expandafter\@slowromancap\romannumeral #1@}

\newcommand{\pan}[1]{\textcolor[rgb]{0,0,0}{#1}}

\def\eg{\emph{e.g.}} 
\def\ie{\emph{i.e.}}

\def\wrt{{w.r.t.~}}

\newcommand{\FullName}{Contrastive Initialization}

\begin{document}

\begin{frontmatter}

\title{Improving Fine-tuning of Self-supervised Models with Contrastive Initialization}

\cortext[equalauthors]{These authors contributed equally to this work.}
\cortext[mycorrespondingauthor]{Corresponding author}

\author[address1]{Haolin~Pan\corref{equalauthors}}
\ead{sephl@mail.scut.edu.cn}
\author[address1]{Yong~Guo\corref{equalauthors}}
\ead{guoyongcs@gmail.com}
\author[address1]{Qinyi~Deng\corref{equalauthors}}
\ead{sedengqy@mail.scut.edu.cn}

\author[address1]{Haomin~Yang}
\ead{sehaomin@mail.scut.edu.cn}

\author[address2]{Yiqun~Chen}
\ead{chenyiqun@gdei.edu.cn}

\author[address1]{Jian~Chen\corref{mycorrespondingauthor}}
\ead{ellachen@scut.edu.cn}

\address[address1]{South China University of Technology, China}
\address[address2]{Guangdong University of Education, China}

\begin{abstract}
Self-supervised learning (SSL) has achieved remarkable performance in pre-training the models 
that can be further used in 
downstream tasks via fine-tuning.
However, these self-supervised models may not capture meaningful semantic information since the images belonging to the same class are always regarded as negative pairs in the contrastive loss. 
Consequently, the images of the same class are often located far away from each other in learned feature space, which would inevitably hamper the fine-tuning process. To address this issue, we seek to provide a better initialization for the self-supervised models by enhancing the semantic information. To this end, we propose a Contrastive Initialization (COIN) method that breaks the standard fine-tuning pipeline by introducing an extra initialization stage before fine-tuning. 
Extensive experiments show that, with the enriched semantics, our COIN significantly outperforms existing methods \emph{{without introducing extra training cost}} and sets new state-of-the-arts on multiple downstream tasks.
\end{abstract}


\end{frontmatter}



\section{Introduction}
\label{Introduction}
Recently, self-supervised learning (SSL) has achieved great success in the pre-training of deep models based on a large-scale unlabelled dataset~\cite{bao2022beit,BISCIONE2022222,chen2021exploring,grill2020bootstrap,he2020momentum}.
Specifically, one can train the models by maximizing the feature similarity between two augmented views of the same instance while minimizing the similarity between two distinct instances. 
In practice, these self-supervised models have shown remarkable generalization ability across diverse downstream tasks when fine-tuning their parameters on the target datasets~\cite{chen2020big,girshick2014rich,hu2021well,huang2021towards,tian2020makes}. 
\pan{To be specific}, existing methods often exploit the cross-entropy loss, optionally combined with a contrastive loss~\cite{tang2021confit,zhang2021few,zhang2021unleashing,zhong2020bi}, to fine-tune the pre-trained models.

\begin{figure*}[t]
\centering
\includegraphics[width=1.0\textwidth]{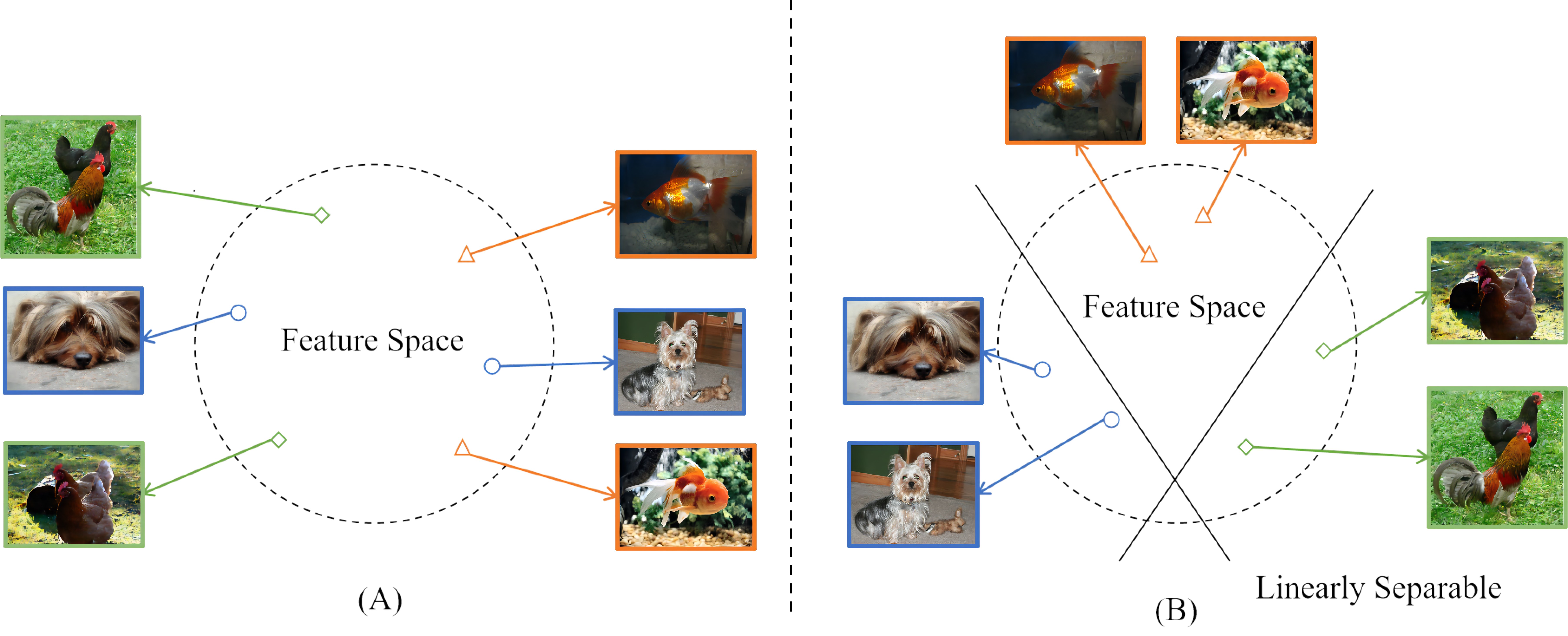}
\vspace{-20pt}
\caption{\label{figure: feature spaces}The feature spaces obtained by self-supervised models. Each mark represents an instance in the feature space, and the shape of the mark indicates the class \pan{of instances}.
Since the similarities of all positive pairs are large enough, and those of all negative pairs are small enough~(\ie, uniformly distributed in the feature space), the values of contrastive loss for both (A) and (B) are small enough to be considered a good feature space. But obviously, the semantic relation among instances in (B) is more meaningful than in (A) due to the separable features. The figure indicates that the goal of self-supervised methods has little positive effect on enriching the semantic information on the downstream tasks. Therefore, it is difficult for the self-supervised models to provide a good initialization for subsequent fine-tuning process.}
\end{figure*}

Nevertheless, the fine-tuning performance is still very limited since the pre-trained model does not necessarily provide a strong/meaningful semantic relation among instances, which, however, is essential for learning a good classifier~\cite{ge2021robust,huynh2022boosting,liu2022enhancing,robinson2021can}.
As shown in Figure~\ref{figure: feature spaces} (A), even on the pre-training dataset, e.g., ImageNet, SSL may learn a feature space where the instances belonging to the same class are far from each other~\cite{wang2021understanding,wang2020understanding}.
The main reason is that SSL only takes different augmented views of the same image as the positive pairs and simply treats all the other images as the negative ones.
As a result, the images belonging to the same category are not necessarily located close to each other in the learned feature space, i.e., with very weak semantic relation. 
More critically, this phenomenon would be much more severe when we consider a target dataset that has a different distribution from the pre-training dataset.
In practice, such a weak semantic relation among instances would inevitably hamper the fine-tuning process from learning a good classifier on the downstream tasks.

To address this issue, when fine-tuning, we seek to provide a better initialization for the self-supervised models by enhancing the semantic relation among instances. 
Intuitively, if we can encourage the model to capture better semantic information on the target dataset, it would be easier to learn a promising classifier during fine-tuning.
To be specific, as shown by the example in Figure~\ref{figure: feature spaces} (B), based on the meaningful semantic relation where the instances 
\pan{belonging to} the same category are close to each other, we can simply use a linear classifier to discriminate the instances of different categories in the feature space.

\begin{figure*}[t]
\centering
\includegraphics[width=1\linewidth]{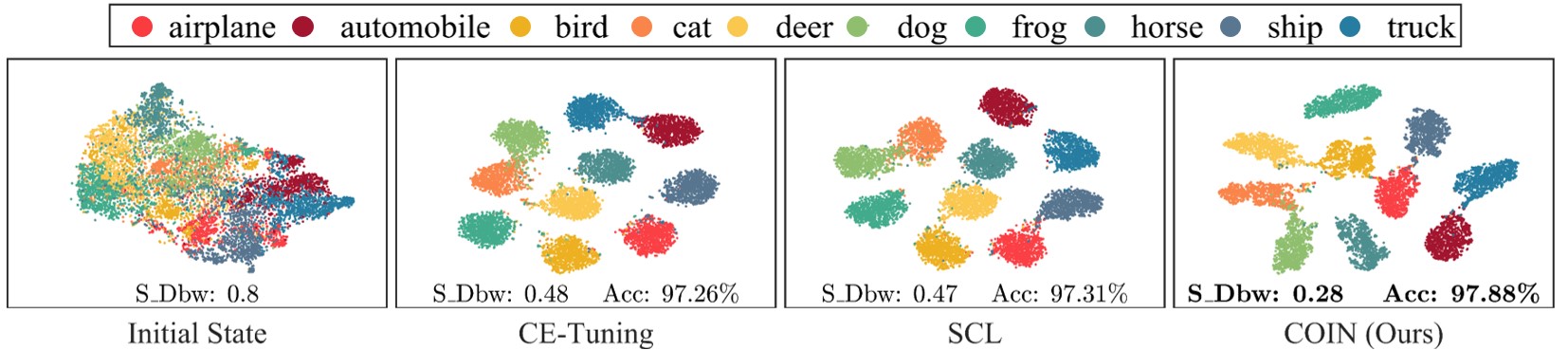} 
\vspace{-20pt}
\caption{
\label{figure: feature visualization}
Visualization of the feature space
on CIFAR-10, with each color representing a specific class.
We compare the initial state and two popular fine-tuning methods, i.e., CE-Tuning with a cross-entropy (CE) loss and SCL~\cite{gunel2020supervised}, which combines CE loss with a supervised contrastive loss.
Clearly, our COIN~effectively pushes away the instances of different classes and yields narrow ellipse areas with fewer overlapping~(\ie, lower $S\_Dbw$ score), coming with better discriminative power.
As a result, COIN~greatly improves the fine-tuning performance.}
\end{figure*}

Inspired by this, we propose to break the pipeline of fine-tuning \pan{self-supervised models} by introducing an extra class-aware initialization stage before fine-tuning the classifier.
In this paper, we develop a \FullName~(COIN) method that exploits a supervised contrastive loss to enrich the semantic information on the target dataset, by pulling together the instances of the same class and pushing away those from different classes.
In this way, we are able to obtain easily separable features with better semantic information and then help learn a good classifier \pan{during fine-tuning}.
To verify this, we visualize the feature space learned by different methods in Figure~\ref{figure: feature visualization}.
Compared with two popular fine-tuning methods, \ie, CE-Tuning~\cite{he2020momentum} and SCL~\cite{gunel2020supervised}, our COIN~effectively pushes away the instances of different classes and yields a narrow ellipse area for each class.
In fact, such narrow ellipse areas often come with better discriminative power, which can be evaluated by the $S\_Dbw$ score (\emph{lower is better}) with the consideration of both the inter-class discrepancy and the intra-class compactness~\cite{Halkidi2001ClusteringVA}.
As shown in Figure~\ref{figure: feature visualization}, our COIN~greatly reduces the $S\_Dbw$ score from $0.48$ to $0.28$ and yields a large accuracy improvement of $>$0.5\% on CIFAR-10.
More critically, we highlight that our initialization stage does not increase the total training cost, since we reduce the number of iterations for the following fine-tuning process to keep the total training iterations unchanged.

Our main contributions can be summarized as follows:
\begin{itemize}

\item
We break the standard fine-tuning pipeline for self-supervised models by introducing an additional  initialization stage before fine-tuning.
To be specific, we first encourage the model to capture better semantic information on the target dataset. Then, we take the model with the enriched
\pan{semantic information} as a better initialization for the \pan{subsequent}
fine-tuning process.
\item 
To enrich the semantic information, we propose a Contrastive Initialization (COIN) method that exploits a supervised contrastive loss to perform class-aware clustering.
Specifically, we pull together the instances of the same class and push away those instances from different classes.
In this way, we can obtain easily separable features with better semantic information on the target dataset, which \pan{significantly}
boost the fine-tuning performance.
\item 
To avoid introducing extra training cost, we reduce the number of training iterations for the final fine-tuning process to keep the total training iterations unchanged.
Extensive experiments show that the proposed method, COIN, significantly improves the fine-tuning process of self-supervised models and yields new state-of-the-arts on various benchmark datasets.
\end{itemize}

\section{Related Work}
\subsection{Contrastive Learning}
In the past few years, Many works applying contrastive learning to pre-train self-supervised models have attracted attention due to their impressive performances~\cite{chen2020simple,chen2020big,chen2020improved,he2020momentum,LIU202299,tian2020makes}.
The self-supervised contrastive learning models learn an instance-distinct-based feature representation to achieve state-of-the-art performance on the ImageNet \cite{deng2009imagenet} classification task.
However, the superior performance of a self-supervised model in one scenario does not necessarily reflect its performance in others.
Because the learned feature spaces closely match the distribution of ImageNet, which easily overfit some similar downstream tasks but hamper others~\cite{chen2021intriguing,kotar2021contrasting,newell2020useful}.
Therefore, in this paper, we focus on better fine-tuning the self-supervised models on various downstream tasks.

\subsection{Model Fine-tuning}
In deep learning, fine-tuning a pre-trained model on a target dataset has become a standard training paradigm in various applications.
Fine-tuning is one of the main pipelines for improving the transferability of self-supervised models. 
To be precise, fine-tuning can be regarded as a model transfer method, but it usually does not need to know the data distribution of the source domain.
Most fine-tuning methods are designed for supervised pre-trained models instead of self-supervised models~\cite{ZHANG20211,BASHA2021112}.
Besides fine-tuning parameters, one can also fine-tune architectures to improve performance~\cite{guo2019nat,guo2021towards}.
In recent studies, how to achieve a better fine-tuning performance of self-supervised models has attracted 
more attention~\cite{li2020rifle,you2020co}, in which contrastive learning plays an important role~\cite{gunel2020supervised,zhang2021unleashing,zhong2020bi}.
Recently, Noisy-Tune~\cite{wu2022noisytune} proposed to improve the fine-tuning performance by introducing noises into the parameters.
Unlike the above methods, we seek to boost the fine-tuning process from a new perspective, i.e., providing a better initialization with strong semantic information.

\subsection{Domain Adaption}
The Purpose of domain adaption is to transfer a trained model from the source domain to the target domain \pan{across the distribution shift}. 
Many domain adaptation studies focus on minimizing the discrepancy between the data distributions of the two domains~\cite{HOU2022172,tian2020makes,guo2020closed,XIE202198,JING202039,guo2022towards}. 
In addition, some studies focus on improving model transferability using source domain data so that the model can be more easily adapted to the target domain~\cite{chen2019transferability,dangovski2021equivariant,xiao2020should}.
Some of them show that self-supervised models often benefit from semantic information and feature uniformity~\cite{wang2021understanding,wang2020understanding,wang2021self,yan2020clusterfit}. 
Besides model adaptation, improving the generalization ability or out-of-distribution robustness~\cite{guo2022improving,schneider2020improving} can also handle the data from another domain to some extent. 
Unlike the above methods that train models on the data of the source domain, we focus on obtaining a better initialization on the target domain/dataset.

\section{Fine-tuning with \FullName}
\subsection{Motivation and Method Overview}
\label{Method: Motivation and Overview}

Since self-supervised learning (SSL) takes different augmented views of an instance as positive pairs and simply treats all other instances as negative pairs, self-supervised models are often hard to capture meaningful semantic information.
As shown in Figure \ref{figure: feature spaces} (A), 
in the feature space learned by SSL, the instances belonging to the same class may be far away from each other, i.e., with weak semantic relation among instances.
In practice, such weak semantic information often hampers the fine-tuning process from learning a good classifier and results in suboptimal fine-tuning performance.

\begin{figure*}[t]
\centering
\includegraphics[width=1\linewidth]{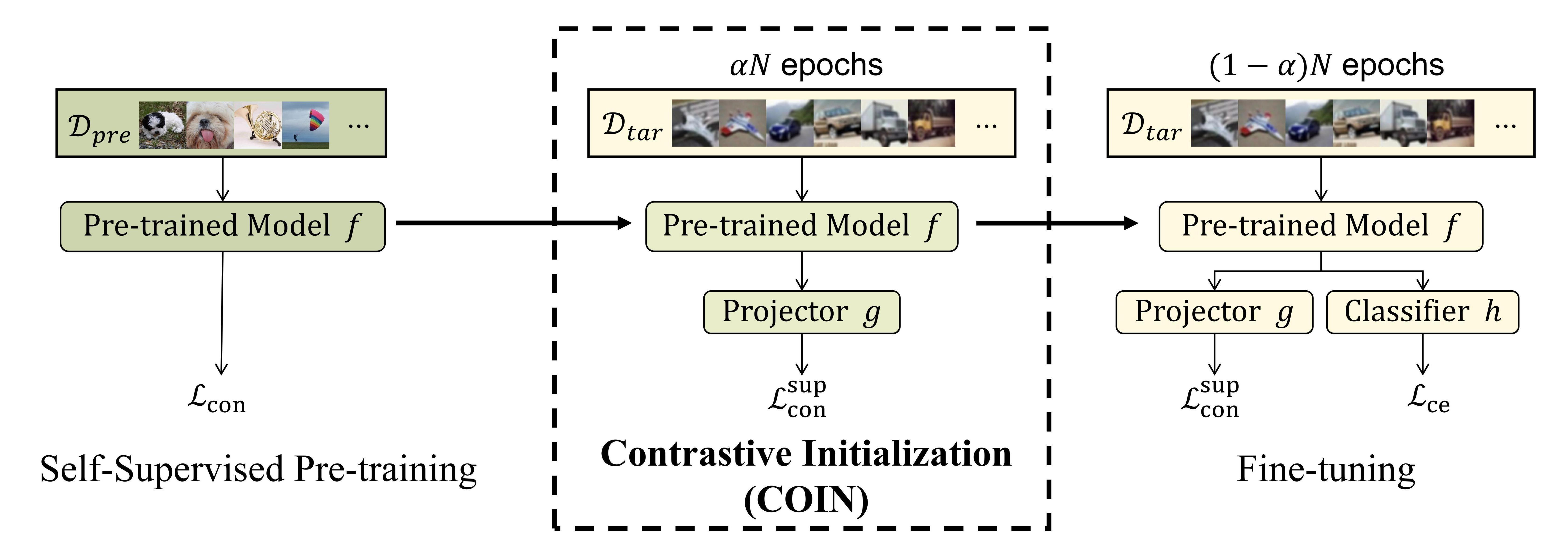}
\vspace{-20 pt}
\caption{\label{figure: overall pipeline}The overall fine-tuning pipeline with COIN. 
We introduce an additional Contrastive Initialization (COIN) stage that exploits a supervised contrastive loss to enrich the semantic information on the target dataset.
Then, we can perform any existing fine-tuning method, \eg, SCL~\cite{gunel2020supervised} with an additional contrastive loss, to learn the classifier.
Besides, we introduce a stage split ratio $\alpha$ to allocate different computational budgets for each stage, while keeping the training epochs $N$ unchanged.}
\end{figure*}

\begin{algorithm}[t!]
\caption{\label{algorithm: COIN}Fine-tuning with \FullName~(COIN).}
\textbf{Require}: Pre-trained model $f$, projector $g$, classifier $h$, step size $\eta$, model parameters $w$, stage split ratio $\alpha$, epochs $N$, weight of contrastive loss $\lambda$.
\vspace{-20pt}
\begin{algorithmic}[1]
\STATE Initialize the classifier $h$ and projector $g$;
\STATE \emph{// \FullName}
\FOR{$i=1$ to $\alpha N$} 
    \STATE Sample a batch of training data $x$ and extract features $z= f(x)$;
    \STATE Compute projected features $g(z)$ to obtain $\mathcal{L}_{\mathrm{con}}^{\mathrm{sup}}$;
    \STATE Update $\{w_f, w_g\}$ by descending the gradient: \\
    \pan{
    ~~~~~~~Update model $f$: $w_f \leftarrow w_f-\eta \nabla_{w_f} \mathcal{L}_{\mathrm{con}}^{\mathrm{sup}}$;\\
    ~~~~~~~Update projector $g$: $w_g \leftarrow w_g-\eta \nabla_{w_g}\mathcal{L}_{\mathrm{con}}^{\mathrm{sup}}$;}
\ENDFOR
    \STATE \emph{// Any Existing Fine-tuning Method, e.g., SCL~\cite{gunel2020supervised}}
    \FOR{$i=\alpha N+1$ to $N$}
    \STATE Sample a batch of training data $x$ and extract features $z= f(x)$;
    \STATE Compute the projected features $g(z)$ to obtain $\mathcal{L}_{\mathrm{con}}^{\mathrm{sup}}$;
    \STATE Compute logits $h(z)$ to obtain $\mathcal{L}_{\mathrm{ce}}$;
    \STATE Update $\{w_f, w_g, w_h\}$ by descending the gradient: \\
    \pan{
    ~~~~~~~Update model $f$: $w_f \leftarrow w_f-\eta \left( \nabla_{w_f} \mathcal{L}_{\mathrm{ce}} + \lambda \nabla_{w_f} \mathcal{L}_{\mathrm{con}}^{\mathrm{sup}} \right)$;\\
    ~~~~~~~Update projector $g$: $w_g \leftarrow w_g-\eta \left( \nabla_{w_g} \mathcal{L}_{\mathrm{ce}} + \lambda \nabla_{w_g} \mathcal{L}_{\mathrm{con}}^{\mathrm{sup}} \right)$;\\
    ~~~~~~~Update classifier $h$: $w_h \leftarrow w_h-\eta \left( \nabla_{w_h} \mathcal{L}_{\mathrm{ce}} + \lambda \nabla_{w_h} \mathcal{L}_{\mathrm{con}}^{\mathrm{sup}} \right)$;}
\ENDFOR
\end{algorithmic}
\end{algorithm}

To address this issue, we seek to provide a better initialization for fine-tuning by enhancing the semantic relation among instances.
Intuitively, as shown in Figure \ref{figure: feature spaces} (B), if we capture better semantic information,
one can obtain an easily separable feature space that makes it easier to learn a good classifier during fine-tuning.
Inspired by this, we break the standard fine-tuning pipeline of self-supervised models by introducing an extra class-aware initialization stage, resulting in a new fine-tuning pipeline that contains two stages.
As shown in Figure \ref{figure: overall pipeline} and Algorithm~\ref{algorithm: COIN}, we first adopt a \FullName~(COIN) stage that exploits a supervised contrastive loss to enrich the semantic information.
Then, we fine-tune the model that contains richer semantic information with any existing fine-tuning approaches. Following~\cite{gunel2020supervised}, besides the cross-entropy loss, we additionally incorporate a contrastive loss since it often benefits the fine-tuning process. 
For fair comparisons with existing fine-tuning methods, we consider COIN~as a part of the overall fine-tuning pipeline and keep the total training iterations unchanged.
Given a budget of $N$ training epochs in total, we introduce a hyperparameter, the stage split ratio $\alpha$, to allocate $\alpha N$ epochs for COIN~and $(1-\alpha) N$ epochs for the subsequent fine-tuning process.

\subsection{\FullName~(COIN)}
\label{Method: COIN}

To provide a better initialization for the subsequent fine-tuning process, we propose a \FullName~(COIN) stage that exploits a supervised contrastive loss to enrich the semantic information. Specifically, on the target dataset, we seek to pull together the instances of the same classes and push away those from different classes.
\pan{With the help of resultant meaningful semantic relation among instances}, our COIN often obtains easily separable features with strong discriminative power. Based on this, we directly take the enriched model as a better initialization to boost the subsequent fine-tuning process.

Let $x$ be the input images.
We extract the features of $x$ through a self-supervised pre-trained model $f$ followed by a projector $g$ via $v = g(f(x))$.
To enhance the semantic information on the target dataset, for the $i$-th sample $x_i$, we take instances of the same class as the positive pairs and those from different classes as the negative pairs.
The training loss of COIN~becomes:
\begin{equation}
\begin{aligned}
    \mathcal{L}_{\mathrm{con}}^{\mathrm{sup}}=-\frac{1}{n}\sum_{i=1}^{n}\frac{1}{\left | P_{i} \right |}\sum_{v_{j}\in P_{i}}^{}
    \log\frac{e^{\left ( v_{i}^{T}v_{j} /\tau \right )}}{\sum_{v_{k} \in A_{i}}e^{\left ( v_{i}^{T}v_{k} /\tau \right )}},
    \label{focal supervised contrastive loss}
\end{aligned}
\end{equation}
where $n$ denotes the number of instances in a mini-batch, $v_{i}$ denotes the features of the $i$-th instance.
$P_i$ and $A_i$ denote the set of positive pairs and all possible pairs \wrt $x_i$, respectively.
To construct data pairs regarding $v_i$ for contrastive learning, we build $A_{i}=\{v_k | k {\leq} n, k {\neq} i\}$ which includes the features of the other instances in this batch excluding $v_i$ itself.
$\tau$ is a temperature coefficient which is important for contrastive loss. 

{\bf Combining COIN with the subsequent fine-tuning process.}
As an initialization method, our COIN~can be easily combined with any existing fine-tuning method. 
In this paper, we combine COIN with \pan{a popular fine-tuning method, SCL~\cite{gunel2020supervised}, which jointly optimizes a CE loss and a supervised contrastive loss.}
The training loss of SCL~\cite{gunel2020supervised} can be formulated as $\mathcal{L} =  \mathcal{L}_{\mathrm{ce}}+\lambda \mathcal{L}_{\mathrm{con}}^{\mathrm{sup}}$. When fine-tuning, we simply introduce a classifier $h$ to compute \pan{the} CE loss and set $\lambda = 0.1$, thus constructing SCL~\cite{gunel2020supervised}.

{\bf Advantages over existing methods.} 
Compared to the existing advanced methods, our COIN significantly boosts the \pan{subsequent} fine-tuning performance on various benchmark datasets. In fact, with the help of the introduced class-aware initialization stage, COIN greatly enriches the semantic information, \pan{which comes with stronger discriminative power} on downstream tasks and then provides a better initialization of self-supervised models for the subsequent fine-tuning. Furthermore, COIN avoids introducing extra training cost to the standard fine-tuning pipeline. Specifically, COIN keeps the total computational budgets unchanged and allocates the budgets to the initialization stage and the fine-tuning stage by a simple hyperparameter $\alpha$.

\section{Experiments}
In this section, we evaluate the effectiveness of COIN on various benchmark datasets, and compare it with other advanced methods.
We will elaborate on the settings and show the experimental results in the following.

\subsection{Datasets and Metrics}
We test on nine datasets \pan{including ImageNet-20~\cite{zhang2021unleashing}, CIFAR-10~\cite{krizhevsky2009learning}, CIFAR-100~\cite{krizhevsky2009learning}, Caltech-101~\cite{fei2004learning}, Stanford Cars~\cite{krause2013collecting}, FGVC Aircraft~\cite{maji2013fine}, Oxford 102 Flowers~\cite{nilsback2008automated}, Oxford-IIIT Pets~\cite{parkhi2012cats}, DTD~\cite{cimpoi2014describing}, covering common image classification tasks including coarse-grained object classification, fine-grained image classification and texture classification.}
We do not directly test on ImageNet but on ImageNet-20 because the model we used is pre-trained on ImageNet.
ImageNet-20 is a subset of ImageNet with 20 classes, including ImageNette and ImageWoof~\cite{Imagenette}. 
Oxford 102 Flowers is obtained from Kaggle.
Caltech-101 is obtained from TensorFlow. Others \pan{are downloaded} from their official websites.

We report the metrics including top-1 accuracy, $S\_Dbw$ score, and training time cost. \cite{wang2021understanding} uses the aggregation degree of similar instances to evaluate the semantic relation among instances, while COIN considers learning higher intra-class compactness and larger inter-class discrepancy by pulling together the instances of the same class and pushing those apart from different classes.
In this way, COIN~can effectively enrich the semantic information on the target dataset.
To quantify the quality of the semantic information, we use $S\_Dbw$ score~\cite{Halkidi2001ClusteringVA} to measure the intra-class discrepancy by average scattering (denoted by $Scat$), and measure the inter-class compactness by inter-class density (denoted by $Dens\_bw$).
Formally, it can be computed by: $S\_Dbw = Scat + Dens\_bw$. A lower $S\_Dbw$ score means richer semantic information, which comes with stronger discriminative power and boosts the fine-tuning performance.

\begin{figure}[t]
\centering
\includegraphics[width=0.7\linewidth]{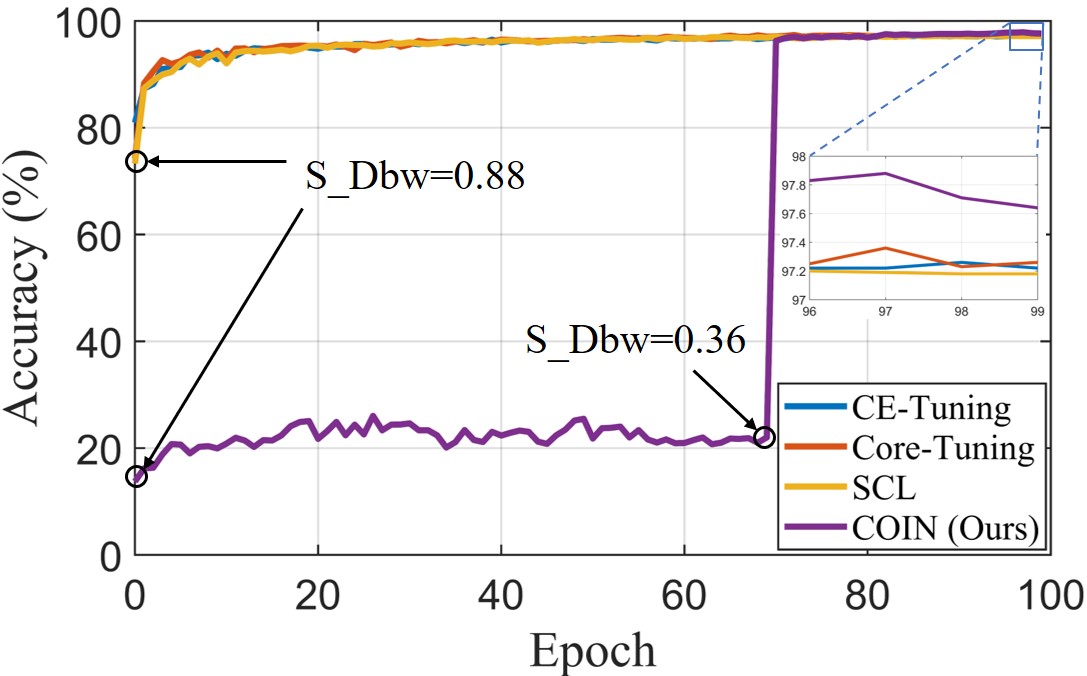} 
\vspace{-5 pt}
\caption{\label{figure: val_acc} Comparisons of the curves of training accuracy among different fine-tuning methods on CIFAR-10. COIN~only uses a supervised contrastive loss during the early 70 epochs, thus showing a lower accuracy, but effectively enriching the semantic information~($S\_Dbw$ score decreases from 0.88 to 0.36) in this period.}

\end{figure}

\subsection{Experimental Settings}
Our experiments follow the nearly same settings as Core-Tuning~\cite{zhang2021unleashing} to fine-tune self-supervised models.
We use ResNet-50 pre-trained by MoCo-v2~\cite{chen2020improved} on ImageNet with 800 training epochs as the pre-trained model, which can be downloaded from the moco GitHub repository.
We implement COIN in PyTorch\footnote{The code of the proposed COIN is available at \href{https://github.com/PorientHaolin/COIN}{https://github.com/PorientHaolin/COIN}.}.
As shown in Algorithm \ref{algorithm: COIN}, we first update the parameters of pre-trained model $f$ and projector $g$ in our COIN. In the subsequent fine-tuning process, besides $f$ and $g$, we also update the parameters of classifier $h$.
For a fair comparison, all methods use consistent data preprocessing schemes on the same \pan{dataset}.

The hyperparameters of our experiments are nearly the same as Core-Tuning \cite{zhang2021unleashing} except for the newly introduced stage split ratio $\alpha$.
Specifically, we set the training epochs $N=100$, the temperature coefficient $\tau=0.3$, and the weight of supervised contrastive loss $\lambda=0.1$. Besides, we set $\alpha=0.7$ on CIFAR-10.
Note that the value of $\alpha$ is specific to the dataset due to the different convergence rates when fine-tuning models on different datasets.
Under the premise of ensuring the self-supervised models in all experiments can converge during the fine-tuning, we slightly adjust the value of $\alpha$ for each dataset to allocate as many computational budgets as possible to the initialization stage.
We finally report the value of $\alpha$ at which the model finally achieves the best accuracy.


\subsection{Comparison with State-of-the-arts}
\label{experiment: comparision}

In this section, we compare COIN~with various advanced fine-tuning methods including CE-Tuning, SL-CE-Tuning~\cite{zhang2021unleashing}, L2SP \cite{xuhong2018explicit}, M\&M~\cite{zhan2018mix}, DELTA~\cite{li2019delta}, BSS~\cite{chen2019catastrophic}, RIFLE~\cite{li2020rifle},  Bi-Tuning~\cite{zhong2020bi}, Noisy-Tune~\cite{wu2022noisytune}, SCL~\cite{gunel2020supervised}, and Core-Tuning~\cite{zhang2021unleashing}. Among them, Noisy-Tune is an initialization method that perturbs parameters before fine-tuning. In like manner of COIN, we \pan{also} combine Noisy-Tune with SCL~\cite{gunel2020supervised} to achieve a fair comparison \pan{with our COIN}.

\begin{table*}[t]
  \centering 
  \caption{\label{tab:acc}Comparisons of accuracy on different classification downstream tasks among various fine-tuning methods with the self-supervised ResNet-50 model pre-trained by MoCo-v2. 
  \pan{COIN significantly boosts the fine-tuning performance and sets new state-of-the-arts on multiple datasets.} * denotes our reproduced results. The last row shows the accuracy improvement of our COIN over SCL~\cite{gunel2020supervised}.}
  \resizebox{\textwidth}{!}{
        \begin{tabular}{l|cccccccccc}
        \toprule
        Method & ImageNet-20 & CIFAR-10 & CIFAR-100 & DTD    & Cars  & Pets  & Flowers & Caltech-101 & Aircraft & Avg. \\
        \midrule
        CE-Tuning    & 88.27 & 97.26 & 82.22 & 69.04  & 89.91 & 91.01 & 99.14 & 92.32 & 88.30 & 88.61 \\
        SL-CE-Tuning~\cite{zhang2021unleashing} & 91.01 & 94.23 & 83.40  & 74.40  & 89.77 & 92.17 & 98.78 & 93.39 & 87.03 & 89.35 \\
        L2SP~\cite{xuhong2018explicit}  & 88.49  & 95.14  & 81.43   & 72.18   & 89.00  & 89.43  & 98.66  & 91.98 & 86.55 & 88.10  \\
        M\&M~\cite{zhan2018mix}  & 88.53  & 95.02  & 80.58   & 72.43   & 88.90  & 89.60  & 98.57  & 92.91 & 87.45 & 88.22  \\
        DELTA~\cite{li2019delta} & 88.35  & 94.76  & 80.39   & 72.23   & 88.73  & 89.54  & 98.65  & 92.19 & 87.05 & 87.99  \\
        BSS~\cite{chen2019catastrophic}   & 88.34  & 94.84  & 80.40   & 72.22   & 88.50  & 89.50  & 98.57  & 91.95 & 87.18 & 87.94  \\
        RIFLE~\cite{li2020rifle} & 89.06  & 94.71  & 80.36   & 72.45   & 89.72  & 90.05  & 98.70  & 91.94 & 87.60 & 88.29   \\
        Bi-Tuning~\cite{zhong2020bi} & 89.06  & 95.12  & 81.42   & 73.53   & 89.41  & 89.90  & 98.57  & 92.83 & 87.39 & 88.58  \\
        
        SCL~\cite{gunel2020supervised}   & 89.29  & 95.33  & 81.49   & 72.73   & 89.37  & 89.71  & 98.65  & 92.84  & 87.44 & 88.54  \\
        SCL*~\cite{gunel2020supervised}   & 89.60 & 97.31 & 82.82 & 73.51 & \textbf{90.32} & 90.91 &   98.90 & 92.74 & 88.33 & 89.38  \\
        Core-Tuning~\cite{zhang2021unleashing} & 92.73  & 97.31  & 84.13  & 75.37   & 90.17  & 92.36  & 99.18  & \textbf{93.46}  & 89.48 & 90.47 \\
        Core-Tuning*~\cite{zhang2021unleashing}  & 93.84 & 97.40 & 83.25 & 74.04 & 90.16 & 92.04 & 98.78 & 92.62 & 88.86 & 90.11  \\
        Noisy-Tune~\cite{wu2022noisytune} & 94.39 & 97.59 & 83.50 & 71.97 & 90.16 & 91.99 & 99.14 & 70.58 & 88.83 & 87.57 \\
        \hline
        \multirow{2}{*}{COIN~(ours)} & \textbf{94.60} & \textbf{97.88} & \textbf{85.39}  & \textbf{75.74}  & \textbf{90.32} & \textbf{93.59} & \textbf{99.27} & 93.00 & \textbf{89.77} & \textbf{91.06} \\
         & ($+5.00$) & ($+0.57$) & ($+2.57$) & ($+2.23$) & ($+0.00$) & ($+2.68$) & ($+0.37$) & ($+0.26$) & ($+1.44$) & ($+1.68$) \\
        \bottomrule
        \end{tabular}
    }
\end{table*}

As shown in Figure \ref{figure: val_acc}, we simply compare COIN~with several representative fine-tuning methods on CIFAR-10. The plotted training curves show how the class-aware initialization stage beneficially affects the subsequent fine-tuning process. We do not \pan{consider} CE loss, but only use a supervised contrastive loss during the early 70 epochs. Therefore, the accuracy of COIN~is lower than other methods in this period.
However, COIN significantly enriches the semantic information on the target dataset~(\ie, decreasing the $S\_Dbw$ score from $0.88$ to $0.36$) before fine-tuning.
With the help of enriched semantic information,
it is easy to learn a more accurate classifier in the following fine-tuning process.

\begin{table*}[t]
\centering
\caption{\label{tab:sdbw}Comparisons of $S\_Dbw$ score and Time Cost of different fine-tuning methods on multiple classification downstream tasks.
$S\_Dbw$ score evaluates the quality of the semantic information captured by models on the target datasets. A lower $S\_Dbw$ score indicates richer semantic information. Time Cost represents the training time for 100 epochs.
These results suggest that COIN  can always obtain the richest semantic information among these methods without introducing additional training cost.}
\resizebox{\textwidth}{!}{
    \begin{tabular}{l|cc|cc|cc|cc|cc}
    \toprule
    \multirow{2}{*}{Method}  & \multicolumn{2}{c|}{ImageNet-20} & \multicolumn{2}{c|}{CIFAR-10} & \multicolumn{2}{c|}{CIFAR-100} & \multicolumn{2}{c|}{Caltech-101} & \multicolumn{2}{c}{DTD} \\
    \cmidrule{2-11}          & $S\_Dbw$ & Time Cost (h) & $S\_Dbw$ & Time Cost (h) & $S\_Dbw$ & Time Cost (h) & $S\_Dbw$ & Time Cost (h) & $S\_Dbw$ & Time Cost (h) \\
    \midrule
    CE-Tuning & 0.75  & 2.23  & 0.48  & \textbf{6.75} & 0.73  & \textbf{7.26}  & 0.32  & \textbf{0.54}  & 0.70  & 0.55 \\
    SCL~\cite{gunel2020supervised} & 0.68  & 2.19  & 0.47  & 7.31  & 0.65  & 7.80  & 0.33  & 0.55  & 0.61  & \textbf{0.51}  \\
    Core-Tuning~\cite{zhang2021unleashing} & 0.55  & 3.72  & 0.38  & 14.93  & 0.64  & 15.64  & 0.32  & 0.61  & 0.62  & 0.76  \\
    COIN~(ours) & \textbf{0.47}  & \textbf{2.10}  & \textbf{0.28} & 7.79  & \textbf{0.46} & 7.79  & \textbf{0.31}  & \textbf{0.54} & \textbf{0.56}  & 0.53  \\
    \midrule
    \multirow{2}{*}{Method} & \multicolumn{2}{c|}{Aircraft} & \multicolumn{2}{c|}{Cars} & \multicolumn{2}{c|}{Pets} & \multicolumn{2}{c|}{Flowers} & \multicolumn{2}{c}{Avg.} \\
    \cmidrule{2-11}          & $S\_Dbw$ & Time Cost (h) & $S\_Dbw$ & Time Cost (h) & $S\_Dbw$ & Time Cost (h) & $S\_Dbw$ & Time Cost (h) & $S\_Dbw$ & Time Cost (h) \\
    \midrule
    CE-Tuning   & 0.43           & 1.54  & 0.49  & \textbf{0.99}  & 0.50  & 0.40  & 0.35  & 0.89  & 0.53  &\textbf{2.35}  \\
    SCL~\cite{gunel2020supervised}  & 0.47           & 1.55  & 0.49  & 1.15  & 0.50  & 0.45  & 0.33  & \textbf{0.84}  & 0.50  & 2.48  \\
    Core-Tuning~\cite{zhang2021unleashing} & 0.45           & 1.76 & 0.51  & 2.05  & 0.50  & 0.81  & 0.34  & 1.58  & 0.48  & 4.65  \\
    COIN~(ours)   & \textbf{0.40}  & \textbf{1.39}  & \textbf{0.45}  & 1.15  & \textbf{0.34}  & \textbf{0.38}  & \textbf{0.30}  & 0.88  & \textbf{0.40}  & 2.50  \\
    \bottomrule
    \end{tabular}
}
\end{table*}

We report more results on the accuracy of all considered methods on multiple benchmark datasets in Table~\ref{tab:acc}.
Under the same settings, COIN outperforms other methods on nearly all considered datasets except Caltech-101 and Cars, but shows comparable performance compared with the best method on these two datasets.
To better illustrate this, we focus on comparing COIN with several representative methods. Among them, CE-Tuning is a standard fine-tuning method, while SCL~\cite{gunel2020supervised} combines CE loss with supervised contrastive loss, which often benefits the fine-tuning process. Core-Tuning and Noisy-Tune are one of the most advanced fine-tuning methods and initialization methods, respectively. For the results in Table~\ref{tab:acc} and Table~\ref{tab:sdbw}, COIN significantly improves SCL~\cite{gunel2020supervised} by simply introducing an extra class-aware initialization stage, and always comes with the most enriched semantic information~(lowest $S\_Dbw$ score). Unlike Core-Tuning, COIN always yields state-of-the-art performance without introducing large training cost. Besides, COIN will not increase the difficulties in fine-tuning like Noisy-Tune which shows limited fine-tuning performance on DTD and Caltech-101 datasets. Extensive results strongly demonstrate the effectiveness of COIN for fine-tuning self-supervised models.


\section{More Discussions}
\label{sec:Discussions}

\subsection{Imapct of Hyperparameters}
In this section, we explore the impact of several important hyperparameters on the resultant fine-tuning performance of COIN.

{\bf Temperature coefficient $\tau$.} 
$\tau$ is an important hyperparameter of contrastive loss. To determine the value of $\tau$ , we consider a candidate set $\tau \in \{0.07, 0.1, 0.3, 0.5, 0.7\}$ like a previous work~\cite{wang2021understanding}. Table~\ref{tab:tau} shows that $\tau= 0.3$ works best under the setting of $\alpha = 0.6$. We empirically set $\tau = 0.3$ in other experiments because no apparent regularity is shown in these results.

\begin{table*}[t!]
\centering
\setlength{\tabcolsep}{4mm}
\caption{\label{tab:tau}Comparisons of the accuracy of COIN~with the different temperature coefficient $\tau$ on CIFAR-10. For these considered candidate values of $\tau$, as the $\tau$ changes, the resultant accuracy does not change with apparent regularity.}
\resizebox{\textwidth}{!}{
\begin{tabular}{l|cccccc}
\toprule
 $\tau$ & 0.07   & 0.1   & 0.3   & 0.5   & 0.7 & Avg.  \\
\midrule
Acc. (\%) & 97.60 & 97.55 & \textbf {97.85} & 97.80 & 96.69  &97.70 \\
\bottomrule
\end{tabular}
}
\end{table*}

\begin{table*}[t!]
\centering
\caption{\label{tab:alpha}Comparisons of the accuracy of COIN~with the different values of stage split ratio $\alpha$ on CIFAR-10. For a specific dataset, a moderate value of $\alpha$~(\eg, $\alpha=0.7$ on CIFAR-10) brings a significant improvement in resultant performance. A too-large value causes the fine-tuning process to lack sufficient time to make the model fully converge and \pan{consequently} leads to limited fine-tuning performance.}
\resizebox{\textwidth}{!}{
\begin{tabular}{c|cccccccccc}
\toprule
 $\alpha$ & 0.1   & 0.2   & 0.3   & 0.4   & 0.5  & 0.6   & 0.7   & 0.8   & 0.9  & Avg.  \\
\midrule
Acc. (\%) & 97.38 & 97.54 & 97.47 & 97.57 & 97.73 & 97.85 & \textbf{97.88} & 97.69 & 96.39  &97.50 \\
\bottomrule
\end{tabular}
}
\end{table*}

{\bf Stage split ratio $\alpha$.} 
To fairly compare with advanced fine-tuning methods, we use $\alpha$ to allocate different training budgets to the initialization stage and the fine-tuning stage, keeping the number of total training iterations unchanged. Specifically, if the entire fine-tuning pipeline lasts $N$ epochs, the initialization stage will last the first $\alpha N$ epochs.
As shown in Table~\ref{tab:alpha}, COIN achieves the best fine-tuning performance on CIFAR-10 when $\alpha=0.7$.
When we set a large value of $\alpha$, fine-tuning stage lacks sufficient time to fully converge the self-supervised models, resulting in a limited fine-tuning performance.
Since the model requires different training iterations to be fully converged when fine-tuning on different datasets, we empirically choose a suitable value of $\alpha$ for each dataset.

\begin{table*}[t!]
\centering
    \caption{\label{tab:epochs}Comparisons of the accuracy of several representative fine-tuning methods with different total training epochs on ImageNet-20. Since the models have almost converged within 100 training epochs, the best accuracy of each method does not improve significantly after increasing the training epochs.}
    \resizebox{\textwidth}{!}{
    \begin{tabular}{l|cccc}
    \toprule
    Epochs & CE-Tuning & SCL~\cite{gunel2020supervised} & Core-Tuning~\cite{zhang2021unleashing} & COIN~(ours) \\
    \midrule
    100 epochs  & 88.27 & 89.60 & 93.84 & \textbf{94.60} \\
    150 epochs  & 88.30 & 89.60 & 93.93 & \textbf{94.80} \\
    \bottomrule
    \end{tabular}
    }
    \vspace{-10pt}
\end{table*}

{\bf Training epochs $N$.}
We follow \cite{zhang2021unleashing} to fine-tune the self-supervised models with 100 training epochs. \pan{In terms of the results shown in Table~\ref{tab:epochs},} after increasing the epochs to 150, all models do not significantly improve because they almost converge within 100 epochs. COIN still maintains the performance gap \pan{over} the \pan{considered} fine-tuning methods.

\begin{table*}[t]
  \centering 
  \caption{\label{tab:backbone}Comparisons of the accuracy of COIN~with different pre-trained models and fine-tuning settings on several benchmark datasets. COIN~achieves higher accuracy when implemented on ViT pre-trained by MAE~\cite{he2021masked}.}
  \resizebox{\textwidth}{!}{
        \begin{tabular}{l|cccccccc}
        \toprule
        Method & ImageNet-20 & CIFAR-10 & CIFAR-100 & DTD   & Cars  & Pets  & Flowers & Avg. \\
        \midrule
        \pan{MoCo-v2-based COIN~\cite{chen2020improved}} & 94.60 & 97.88 & 85.39  & 75.74  & 90.32 & 93.59 & 99.27 & 90.97 \\
        \pan{MAE-based COIN~\cite{he2021masked}} & \textbf{95.90} & \textbf{98.27} & \textbf{85.62}  & \textbf{76.12}  & \textbf{90.80} & \textbf{94.61} & \textbf{99.63} & \textbf{91.56} \\
        \bottomrule
        \end{tabular}
    }
\end{table*}

\subsection{Applying COIN~to Vision Transformer}
The previous experimental results fully demonstrate the effectiveness of our COIN on ResNet-50. In this section, we simply apply our COIN to a more advanced architecture and then fine-tune on those datasets whose input size is $224 \times 224$, using different fine-tuning settings. Specifically, we use Vision Transformer~(ViT) pre-trained by AutoEncoder~(MAE)~\cite{he2021masked} as the pre-trained model , and follow the fine-tuning settings obtained from the \pan{Github repository of MAE which is implemented on PyTorch}. As shown in Table~\ref{tab:backbone}, with a stronger pre-trained model, our COIN obtains better results and sets new state-of-the-arts across different datasets.

\section{Conclusion}
This paper studies how to better fine-tune self-supervised models on various downstream tasks. We observe that the contrastive loss of SSL treats the instances belonging to the same class strictly as negative pairs, resulting in the limited capability of the self-supervised models to capture meaningful semantic information. We argue that the resultant weak semantic relation among instances indeed hampers the subsequent fine-tuning process. 
In response to this challenge, we thus propose a Contrastive Initialization (COIN) method to provide a better initialization of self-supervised models by simply introducing an extra class-aware stage before fine-tuning. As a result, COIN breaks the standard fine-tuning pipeline, leading to a new one.
Specifically, COIN exploits a supervised contrastive loss to enrich the semantic information, which comes with stronger discriminative power and significantly improves the fine-tuning performance.
We compare COIN with the advanced fine-tuning methods and initialization methods according to multiple evaluation metrics. The promising experimental results on various benchmark datasets show that COIN significantly benefits the subsequent fine-tuning process and consequently yields state-of-the-art performance without introducing additional training cost.




\bibliography{ref}





\end{document}